\documentclass{article}

\usepackage[preprint]{corl_2026}

\usepackage[utf8]{inputenc}
\usepackage[T1]{fontenc}
\usepackage{footmisc}

\usepackage{url}
\usepackage{booktabs}
\usepackage{multirow}
\usepackage{amsmath}
\usepackage{amsfonts}
\usepackage{amssymb}
\usepackage{nicefrac}
\usepackage{microtype}
\usepackage{graphicx}
\usepackage{subcaption}
\usepackage[framemethod=TikZ]{mdframed}
\usepackage{enumitem}
\usepackage{xcolor}
\usepackage{colortbl}
\usepackage{array}
\usepackage{tabularx}
\usepackage{cleveref}
\newcolumntype{Y}{>{\centering\arraybackslash}X}
\definecolor{mainrow}{gray}{0.92}

\newcommand{\pmsd}[1]{{\scriptsize$\pm$#1}}

\newmdenv[
	linewidth=0.5pt,
	linecolor=black!40,
	backgroundcolor=black!3,
	roundcorner=3pt,
	innertopmargin=8pt,
	innerbottommargin=8pt,
	innerleftmargin=10pt,
	innerrightmargin=10pt,
	skipabove=\topsep,
	skipbelow=\topsep,
]{statementbox}

\hypersetup{pdfauthor={Yoo Sung Jang and Michael S. Ryoo},
            pdftitle={Latent Projection for Latent Planning}}

\title{Latent Projection for Latent Planning}

\author{
	Yoo Sung Jang \\
	Stony Brook University \\
	\And
	Michael S.~Ryoo \\
	Stony Brook University \\
}

\begin{document}
\maketitle

\begin{abstract}
Latent world models let an agent plan in latent space. The planner searches
for an action sequence that minimizes a cost, and that cost is often the
$\ell_2$ distance between the predicted terminal latent and the goal latent.
The latent, however, is partly noise: the regularizer that prevents collapse
during world model training promotes variance in the directions carrying no
signal. We show that the
same regularizer also supplies the structure to prune that noise in the
latent. SIGReg drives the latent marginal to a standard Gaussian at every
step, and assuming further that the latent sequence is a Gauss--Markov process, the
transition is linear with its covariance
pinned by the map itself, $z' \mid z \sim \mathcal{N}(Az,\, I - AA^\top)$.
Fitting the state dynamics $A$ therefore yields the covariance at no
additional cost, and in its eigenbasis each latent direction carries its own
uncertainty. Furthermore, constraining $A$ to be a normal matrix fixes that
basis across timesteps, so one eigenbasis measures the uncertainty over an
arbitrary number of steps and serves as a stable basis to compute the
planning cost on. We then restrict the $\ell_2$ cost to the $k$-dimensional
subspace the state dynamics carry forward with the least uncertainty. The correction is
post hoc: the encoder, the predictor, the rollout, and the planner are all
left untouched, and only a single constrained matrix, $A$, is fit. We
evaluate across six environments and show large gains on
the 2D navigation tasks, TwoRoom and PointMaze, raising success rate over
the plain $\ell_2$ cost by up to $+54.7$ points under a harder planning
setting.
Code is available at \url{https://github.com/yjang43/lp2}.
\end{abstract}


\section{Introduction}
\label{sec:intro}

Latent world models let an agent plan in representation space. An encoder
compresses an observation into a latent representation, a predictor advances that latent
under an action, and a planner searches for an action sequence that carries the
latent close to the goal. Every part of the loop involves the latent,
and learning this representation is a central design choice
that varies across latent world models
\citep{maes2026leworldmodel, assran2025v, zhou2024dinowm, hansen2024td,
hafner2025mastering}.

Joint-Embedding Predictive Architectures (JEPAs)
\citep{lecun2022path}, a recipe for general representation learning, are
widely chosen for their simplicity: a JEPA predicts the target's representation
from the context's representation, and that prediction objective is what shapes
the representation. Left unconstrained, the representation collapses to a constant,
so JEPAs rely on a regularizer to keep the
embedding informative. Recent instances of JEPA use regularizers that encourage
the embedding to lie in a structured distribution, most commonly an isotropic
Gaussian or a variant of it
\citep{balestriero2025lejepa, wu2026visreg, kuang2026rectified}.

LeWorldModel (LeWM) \citep{maes2026leworldmodel} builds a latent world model on this recipe,
training an encoder and predictor end to end from pixels with a two-term loss:
next-latent prediction and SIGReg \citep{balestriero2025lejepa}, which pulls the
embedding toward an isotropic Gaussian. Similar to many other latent world models
\citep{zhou2024dinowm, assran2025v, sobal2025stress},
LeWM plans by ranking candidate action sequences on the $\ell_2$ distance
between the predicted terminal latent and the goal latent.

The simplicity of the cost, however, overlooks two sources of noise in the representation:
(i) the regularizer promotes variance in which noise fills the directions
carrying no signal; and (ii) the encoder, trained to represent observations,
retains content irrelevant to planning
\citep{zhao2026sub, wang2026temporal, li2026beyond}. The distance is then
measured partly along directions that carry no information about reaching the goal.

Ironically, the same regularizer that injects the noise also gives the
structure to repair it. The embedding is driven to be stationary under the
transition, and reading the latents as a Gaussian world model in the sense of
\citet{klindt2026does}, we find the noise is not free but pinned to
the transition itself: fit the state dynamics and the covariance comes with it.
By projecting onto a subspace determined by the covariance, we obtain a
representation restricted to the directions that are certain with respect to
the state dynamics. In general, the subspace moves under the transition,
but we show that it is fixed under a normality constraint on the state
dynamics. All of
this is post hoc: the only thing we fit is a constrained linear function; the
encoder, the predictor, and the planner are untouched.

\noindent Our main contributions are:
\begin{enumerate}[leftmargin=*, topsep=4pt, itemsep=3pt]
	\item \textbf{Uncertainty measure in a time-invariant basis.}
	Under SIGReg's standard Gaussian marginal and a Gauss--Markov model,
	the noise is pinned to the dynamics,
	$z' \mid z \sim \mathcal{N}(Az,\, I - AA^\top)$, where each eigenvector
	carries its own uncertainty. Each step of the state dynamics generally
	moves that basis. We show that when $A$ is a normal matrix, the retained subspace
	is closed under the dynamics and the same basis serves every timestep.

	\item \textbf{A new planning cost.} We measure the cost on the subspace
	spanned by the $k$ eigenvectors of $I - AA^\top$ with the smallest
	eigenvalues: the directions the state dynamics carry forward with certainty.
	Empirically, planning with this cost delivers large gains in success
	rate on
	the 2D navigation tasks, TwoRoom and PointMaze, of up to $+54.7$
	points under a harder planning setting.
\end{enumerate}

\section{Preliminaries: LeWorldModel}
\label{sec:prelim}
\label{subsec:lewm}

LeWM \citep{maes2026leworldmodel} is a latent world model that learns the
representation and the state-action dynamics together. An encoder $f$ maps an
observation $x$ to a latent $z = f(x) \in \mathbb{R}^n$, and a predictor
$g$ outputs the future latent, $\hat{z}' = g(z, a)$, under an action $a$.

\paragraph{Training.} The learning is driven by two terms, the predictor
loss and the SIGReg loss \citep{balestriero2025lejepa}:
\begin{equation}
	\min_{f, g} \; \mathbb{E}\big[\, \big\| g(f(x), a) - f(x') \big\|^2 \,\big]
	+ w \, \mathcal{R}_{\mathrm{SIGReg}} .
	\label{eq:lewmloss}
\end{equation}
The SIGReg term pulls the latent distribution toward $\mathcal{N}(0, I)$, which
prevents representation collapse; it acts on the encoded latents at every step of the
trajectory, so $z$ and $z'$ are driven toward the same marginal, providing stationarity in transition.

\paragraph{Planning.}
Given an initial observation $x_0$ and a goal observation $x_{g}$,
a candidate action sequence $a_{1:T}$ is rolled out from $z_0 = f(x_0)$
through the predictor, and its terminal latent $\hat{z}_T$ is scored by the
distance to the goal latent
$z_{g} = f(x_{g})$:
\begin{equation}
	\mathcal{C}(a_{1:T}) = \big\| \hat{z}_T - z_{g} \big\|_2 .
	\label{eq:l2cost}
\end{equation}
A sampling-based planner \citep{rubinstein2004cross} ranks candidates by this
cost and iteratively refines its sampling distribution toward better action
sequences.
The cost, however, falls short as a planning signal:
distance in the latent space does not measure progress on the task, the
regularizer injects noise into the embedding, and the representation also
holds details irrelevant to it. This weakness is
well recognized
\citep{wang2026temporal, bai2026temporal, li2026predictive,
thil2026subspace, zhao2026sub}.

In this work, we take the encoder and predictor as given: both stay
frozen throughout, and we modify only the cost \eqref{eq:l2cost}, replacing
the plain $\ell_2$ distance with one computed after projecting latents onto
a subspace that reflects task-relevant dynamics.

\section{Transition under a Gaussian World Model}
\label{sec:transition}

Our objective is to identify the latent subspace that retains clean
\emph{state dynamics}. In this section, we model the latent space such that
the state dynamics take the form of a normal matrix, and show that its
covariance is pinned by the transition itself.
This covariance in turn identifies the subspace: the directions the state
dynamics carry forward with certainty.

This follows from Gaussian world model assumptions similar to those of
\citet{klindt2026does}.
By construction of LeWM training, the latent marginals are
standard Gaussian and stationary: SIGReg drives the encoded latent toward
$\mathcal{N}(0, I)$ at every step of a trajectory. The rest is our modeling
choice: we assume the latent sequence is a Gauss--Markov process, and
further assume the state dynamics act independently in some orthonormal
basis (on lines and planes).

\begin{statementbox}
\textbf{Model (state dynamics under a Gaussian world model).}
\begin{enumerate}[label=(\roman*), leftmargin=*, topsep=4pt, itemsep=4pt]
	\item \textbf{Markov.} At every step,
	$p(z' \mid z, \ldots) = p(z' \mid z)$.
	\item \textbf{Joint Gaussianity.} The pair $(z, z')$ is jointly
	Gaussian.
	\item \textbf{Stationarity.} Both marginals are standard Gaussian:
	$z, z' \sim \mathcal{N}(0, I)$.
	\item \textbf{Normality.} The state dynamics act independently on
	orthogonal lines and planes (see \Cref{subsec:normality}).
\end{enumerate}
\end{statementbox}

\citet{klindt2026does} prove that LeJEPA-style training
\citep{balestriero2025lejepa} recovers the true latent of a Gaussian world
model up to a rotation;
\citet{zhang2026identifiability} extend this to an action-conditioned
predictor as the learner. Every assumption above is invariant under
rotation, so we read the model as a statement about the true latent, not
about a particular embedding.

\subsection{The Transition is Linear with Pinned Noise}
\label{subsec:linear}

By the Markov property (i) and joint Gaussianity (ii), the future latent is
expressed as $z' = m(z) + \eta$ at every step, where the state dynamics
$m(z)$ are affine and the noise $\eta$ is Gaussian and independent of $z$.
Stationarity (iii) centers both marginals, so $m(z) = Az$, and pins the
noise covariance to keep the marginals at $\mathcal{N}(0, I)$. As in
\citet{klindt2026does}, we have an Ornstein--Uhlenbeck transition, now
determined by a single matrix $A$:
\begin{equation}
	\boxed{\;
		z' \mid z \;\sim\; \mathcal{N}\!\big(Az,\; I - AA^\top\big)
	\;}
	\label{eq:main}
\end{equation}
Whatever variance is contracted under the
transition is refilled with exactly that much noise, keeping the marginal
standard Gaussian. Two consequences follow. First,
the noise is pinned to the transition: fit $A$, and
$\Sigma = I - AA^\top$ comes for free. Second, since
$\Sigma \succeq 0$, every singular value of $A$ lies in $[0, 1]$.

The covariance further tells us where the uncertainty lies. Since the
covariance is symmetric, its eigenvectors form an orthonormal basis:
$I - AA^\top = U \operatorname{diag}(\sigma_1^2, \ldots, \sigma_n^2) U^\top$.
Rotating the latent into that basis, $\bar{z} = U^\top z$, leaves the
transition in the same form,
\begin{equation}
	\bar{z}' \mid \bar{z} \;\sim\;
	\mathcal{N}\!\big(\bar{A} \bar{z},\; I - \bar{A}\bar{A}^\top\big) ,
	\label{eq:eignoise}
\end{equation}
where $\bar{A} = U^\top A U$ and the covariance is diagonal,
$I - \bar{A}\bar{A}^\top
= \operatorname{diag}(\sigma_1^2, \ldots, \sigma_n^2)$. \emph{Along this
basis the noise is independent, and each direction $u_i$ carries its own
uncertainty $\sigma_i^2$.} Retaining those of smallest $\sigma_i^2$ isolates
a subspace with clean state dynamics.

\subsection{Normality of the State Dynamics}
\label{subsec:normality}

The remaining assumption (iv) states that the state dynamics are
\emph{normal}: the matrix $A$ satisfies
\begin{equation}
	\boxed{\;
		A A^\top = A^\top A
	\;}
	\label{eq:normal}
\end{equation}
In the spirit of the independence assumption in \citet{klindt2026does},
we assume the state dynamics evolve in some basis in an independent
manner. A real normal matrix decomposes as $A = Q B Q^\top$
\citep{horn2012matrix}, with $Q$ orthogonal and $B$ block diagonal:
$1 \times 1$ blocks $\lambda$, each scaling a line, and $2 \times 2$ blocks
$r R(\theta)$, each rotating a plane by $\theta$ and scaling it by $r$.\footnote{$r R(\theta) = r \left(\begin{smallmatrix} \cos\theta & -\sin\theta \\ \sin\theta & \cos\theta \end{smallmatrix}\right)$, a rotation by $\theta$ scaled by $r \ge 0$.} In the rotated coordinates,
$\bar{z} = Q^\top z$, the transition becomes
$\bar{z}' \mid \bar{z} \sim
\mathcal{N}\big(B \bar{z},\, I - B B^\top\big)$,
where the mean and the noise are both block diagonal. Unlike in
\Cref{subsec:linear}, the transition is block-wise independent: each block is
a transition of its own, carrying uncertainty $1 - \lambda^2$ on a line and
$1 - r^2$ on a plane. Note that we may take $Q = U$.

Beyond the independence itself, normality buys us two things.

\paragraph{Shared basis across every horizon.} For any $A$, iterating
\eqref{eq:main} gives the $t$-step transition,
\begin{equation}
	z^{(t)} \mid z \;\sim\;
	\mathcal{N}\!\big(A^t z,\; I - A^t (A^t)^\top\big),
	\label{eq:multistep}
\end{equation}
but for a non-normal $A$ the eigenbasis of that covariance generally
changes with $t$, so the predictable subspace depends on the horizon.\footnote{With $A = U S V^\top$, $A^2 = U S (V^\top U) S V^\top$, and that middle $V^\top U$ leaves $I - A^2 (A^2)^\top$ no longer diagonal in $U$.} Normality
provides the equality $A^t (A^t)^\top = (A A^\top)^t$, under which the
basis $U$ of \eqref{eq:eignoise} is shared across every horizon. A transition
fitted at a different frame skip should then share the basis and differ
only in spectrum. Along the basis, the uncertainty grows as
$1 - \lambda^{2t}$ on a line and $1 - r^{2t}$ on a plane.
What motivates a horizon-independent basis is discussed further in
\Cref{app:whynormal}.

\begin{figure}[t]
	\centering
	\includegraphics[width=\textwidth]{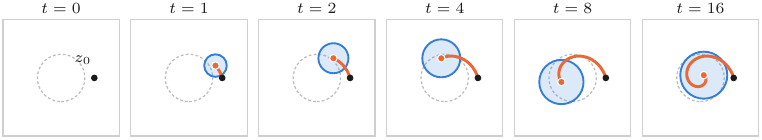}
	\caption{How uncertainty grows across transitions in a $2 \times 2$ block
	--- one independent plane, given by normality. Starting from $z_0$ (black),
	the state dynamics rotate by $\theta$ per step while
	contracting by $r$ (orange). The blue circle marks the uncertainty the
	transition leaves, drawn at one standard deviation; it widens toward the
	unit circle of the stationary law $\mathcal{N}(0, I)$ (grey, dashed).}
	\label{fig:ouspiral}
\end{figure}

\paragraph{Static and motion content, separated.} The blocks distinguish
two kinds of predictable content. A $1 \times 1$ block predicts
\emph{static} content: $\lambda$ fixes the uncertainty $1 - \lambda^2$ and
nothing more, so all it captures is how much of the feature survives the
transition. A $2 \times 2$ block predicts \emph{motion} content: $r$ fixes
the uncertainty $1 - r^2$ the same way, and moreover $\theta$ captures how
the feature moves in the plane under the transition; at $\theta = 0$ the
feature stays put, so the block reduces to predicting static content. \Cref{fig:ouspiral} illustrates
a motion block across transitions.

How much of the latent is motion versus static content is set by a
hyperparameter: $P$ planes and $L = n - 2P$ lines, where $n$ is the latent
dimension. We use the all-planar layout,
$P = \lfloor n/2 \rfloor$, unless otherwise stated.

\section{Latent Projection for Latent Planning}
\label{sec:projection}

We now turn \eqref{eq:main} into a planning cost. The method has two stages:
fit the state dynamics $A$ to trajectories encoded by a pretrained encoder,
then use
the covariance it determines to decide which latent directions the planner
should compare. LeWM's encoder and predictor stay frozen; planning then runs
on them as before, with $A$ shaping the cost.

\subsection{Fitting the State Dynamics}
\label{subsec:fit}

We encode the trajectory data once with the frozen encoder $f$, and fit $A$
to the latent transitions by least squares,
\begin{equation}
	\min_{A} \; \mathbb{E}\big[\, \| A z - z' \|^2 \,\big] .
	\label{eq:fitA}
\end{equation}
Note that the fit is action-free, so $A$ describes the state dynamics under the
policy that generated the data.
The fit must respect two constraints: $\Sigma = I - AA^\top$ must be a
valid covariance, $\Sigma \succeq 0$, and $A$ must be a normal matrix, as discussed in
\Cref{sec:transition}. We parameterize $A$ so that both hold by
construction; \Cref{app:param} gives the full parameterization.

\begin{figure}[t]
	\centering
	\begin{subfigure}[b]{0.44\textwidth}
		\centering
		\includegraphics[width=\textwidth]%
		{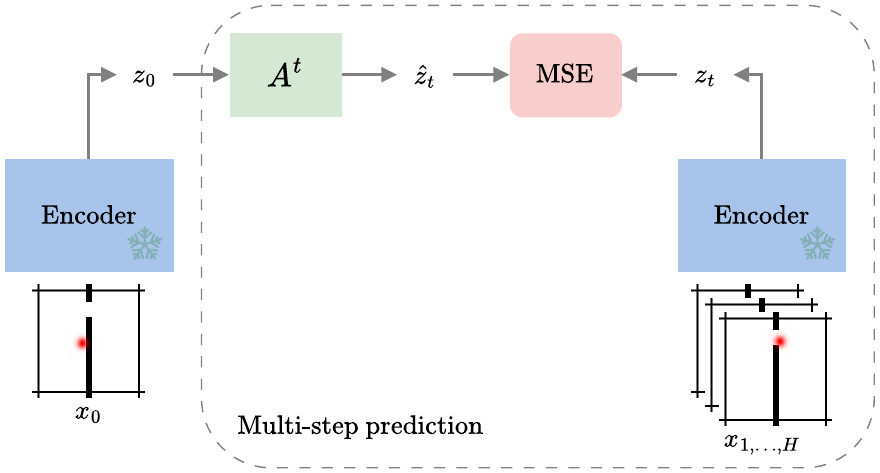}
		\caption{Multi-step fitting.}
		\label{fig:multistep}
	\end{subfigure}
	\hfill
	\begin{subfigure}[b]{0.54\textwidth}
		\centering
		\includegraphics[width=\textwidth]%
		{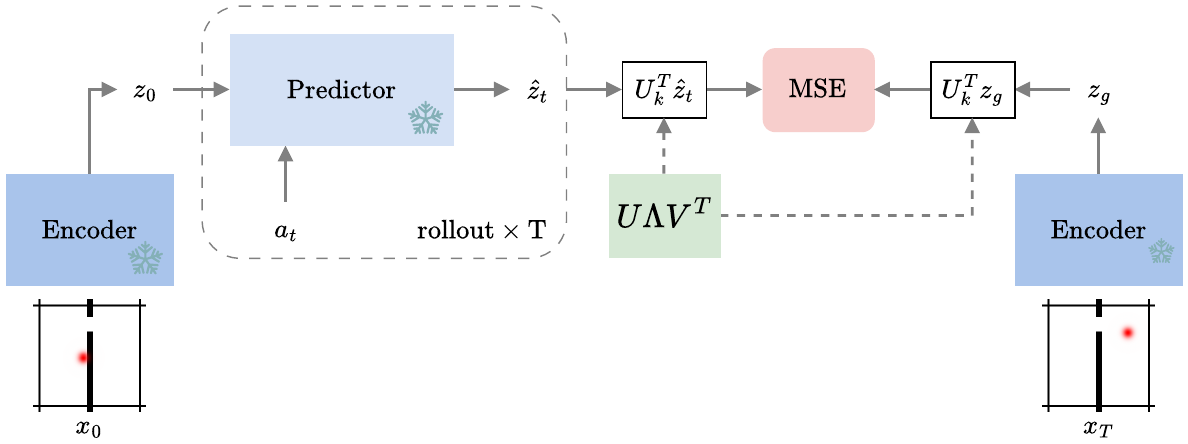}
		\caption{Latent projection for latent planning.}
		\label{fig:planning}
	\end{subfigure}
	\caption{(a) The frozen encoder maps the initial observation and each of
	the $H$ future observations to latents, and $A^t$ is fit to the $t$-step
	transition by squared error. $A$ is the only trained component.
	(b) A candidate action sequence is rolled out through the frozen
	predictor, and its terminal latent is compared to the goal after both are
	projected onto the retained directions $U_k$.}
	\label{fig:method}
\end{figure}

\paragraph{Multi-step fitting.} The $t$-step transition is just $A^t$
\eqref{eq:multistep}, and trajectory datasets
supply $H$ future frames. We therefore fit multiple
timesteps jointly,
\begin{equation}
	\min_{A} \;
	\sum_{t=1}^{H} \mathbb{E}\big[\, \| A^t z - z^{(t)} \|^2 \,\big],
	\label{eq:multistepfit}
\end{equation}
so that a single $A$ has to account for the $t$-step transition at every
$t \le H$ (\Cref{fig:multistep}).

\subsection{Planning in the Projected Latent}
\label{subsec:project}

Ranking action sequences by latent distance is a statement about
reachability, and reachability is governed by the state dynamics. Fitting $A$
recovers those dynamics and their uncertainty measure. Restricting the cost to the
certain directions can then sharpen the ranking.

Fitting $A$ sets the covariance as $\Sigma = I - AA^\top$ at no additional
cost. Its eigendecomposition
\begin{equation}
	\Sigma = U \, \mathrm{diag}(\sigma_1^2, \ldots, \sigma_n^2) \, U^\top,
	\qquad \sigma_1 \le \cdots \le \sigma_n ,
	\label{eq:eig}
\end{equation}
gives the basis that the cost will use: the $i$-th column $u_i$ of $U$ carries
noise variance $\sigma_i^2$, and the columns are ordered from most to least
certain.

We keep the $k$ most certain directions. Writing
$U_k \in \mathbb{R}^{n \times k}$ for the corresponding columns, we replace
the plain $\ell_2$ cost \eqref{eq:l2cost} with the distance
computed after projecting both latents onto that subspace,
\begin{equation}
	\mathcal{C}_k(a_{1:T})
	= \big\| U_k^\top \hat{z}_T - U_k^\top z_{g} \big\|_2 ,
	\label{eq:projcost}
\end{equation}
where $\hat{z}_T$ is obtained by rolling the candidate action sequence forward
through the frozen predictor $g$ exactly as before. Only the cost changes; the
encoder, the predictor, the rollout, and the planner are all untouched.
Taking $k = n$ recovers the plain $\ell_2$ cost, since the projection is then
a rotation, which leaves distances unchanged.

\paragraph{The subspace does not depend on the horizon.}
The $t$-step transition \eqref{eq:multistep} carries noise
$I - A^t(A^t)^\top$. By the normality assumption this equals
$I - (AA^\top)^t$, which shares the eigenvectors of $\Sigma$ at every $t$
(\Cref{subsec:normality}).
Only the eigenvalues change, growing monotonically with $t$. One
basis therefore serves every horizon. For a non-normal $A$ the basis
generally changes with $t$, and no single choice serves every horizon. This is what motivates
building normality into the parameterization of $A$.

\section{Experiments}
\label{sec:experiments}

The experiments answer three questions: whether the cost under the projected
latent representation improves planning and how it compares to other
baseline methods (\Cref{subsec:mainexp,subsec:baselines}); whether the
effect relies on the structure the regularizer imposes on the embedding
(\Cref{subsec:regularizer}); and whether the structure imposed on the state
dynamics matters (\Cref{subsec:ablation}).

\subsection{Environments and Planning Setup}
\label{subsec:setup}

We evaluate on six environments: TwoRoom, PushT, OGBench-Cube, Reacher,
PointMaze-UMaze, and PointMaze-Medium. For each we take a pretrained
checkpoint and keep the encoder and predictor frozen, with per-task details
deferred to \Cref{app:envs}. The state dynamics $A$ are fit under the
parameterization of \Cref{app:param}, on the same trajectory data that trained
the world model, at frame skip $5$ and prediction horizon $H = 4$, for $50$
epochs with AdamW at learning rate $10^{-3}$ and batch size $1024$. The latent
dimension is $n = 192$ throughout, and the rank of the projection is swept over
$k \in \{2, 4, 8, \ldots, 128, 192\}$ on validation seed $0$ to pick the best
rank $k^\star$. Taking $k = n$ recovers the plain $\ell_2$ cost that LeWM plans
with.
We fit $A$ with $3$ seeds $\{0, 1, 2\}$ and report the mean
and standard deviation of planning success rate. Evaluation is held fixed at
test seed $1$, so every entry is scored on the same $50$ initial states and
goals, matching the evaluation protocol of prior work
\citep{maes2026leworldmodel, zhou2024dinowm}.

We adopt the planning configuration of LeWM throughout: a cross-entropy method (CEM) solver
\citep{rubinstein2004cross} with
a planning horizon of $5$ steps, our \emph{default} setting; the exact
configuration is deferred to
\Cref{app:planning}. We also evaluate a \emph{harder} setting with planning
horizon $10$, where the choice of cost has a larger effect on success rate.

\subsection{Planning under Projection}
\label{subsec:mainexp}

To answer whether the cost under the projected latent representation
improves planning, we plan with $\mathcal{C}_k$ \eqref{eq:projcost} at every rank of the
sweep. As a control, we project the latent onto a random orthonormal basis
of rank $k$ instead, resampled for each seed. Any gap between
the two comes from keeping the informative latent directions. At $k = n$ both recover the
plain $\ell_2$ cost \eqref{eq:l2cost} exactly, since projection onto a full
orthonormal basis preserves the distance.

\begin{figure}[t]
	\centering
	\includegraphics[width=\textwidth]{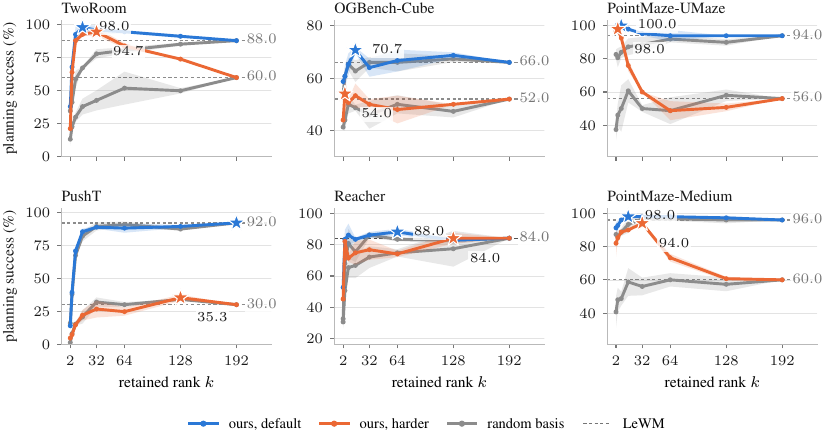}
	\caption{Planning success rate (\%) by retained rank $k$ on validation
	seed, mean $\pm$ std over $3$ seeds of $50$ episodes. Every task is run in
	both the default and harder settings.
	A random basis at the same retained rank is included as a control,
	drawn in gray. At
	$k = 192$ every basis recovers LeWM, drawn as the dashed rule; a star
	marks the selected rank $k^\star$.}
	\label{fig:ranksweep}
\end{figure}

\Cref{fig:ranksweep} reports the sweep on the validation seed. On most
tasks, a small
number of ranks already matches what the cost computed on the full basis
achieves, and projection onto a random basis behaves similarly. This
indicates the latent representation carries a great deal of information
irrelevant to planning. \Cref{app:spectrum} supports this, reporting the
eigenvalues of the transition noise, where the state dynamics are typically
uncertain along a large part of the basis. More subtly, our cost matches the
full-basis success rate at a lower rank than the random control. The same
ordering appears in state recovery (\Cref{app:staterec}), where a linear probe recovers the ground-truth state from the
retained directions well before a random subspace of the same size does.

We now turn to where the projection helps. It wins clearly on
TwoRoom, PointMaze-UMaze and PointMaze-Medium. The gain is most significant
under the harder setting.
Projection onto a random basis shows no comparable gain at any rank.
On the remaining tasks the gain is marginal: PushT needs the full basis, and
Reacher and OGBench-Cube show no such gain over the random control.
Planning can fail in three places: the information the task
requires may be absent from the representation, the predictor may not advance
the latent accurately over the horizon, or the cost may not capture
closeness to the goal latent. Our method repairs only the cost, cannot
compensate for a failure of the other two, and rests on the model of
\Cref{sec:transition}, whose assumptions may not hold under contact-rich
dynamics.

\subsection{Comparison to Other Planning Costs}
\label{subsec:baselines}

Our method projects the latent onto the subspace the state dynamics
carry forward with certainty. We compare it against other planning cost
baselines that run on the same frozen world model:
\begin{itemize}[leftmargin=*, topsep=2pt, itemsep=2pt]
	\item \textbf{TD-JEPA (frozen)} \citep{bai2026temporal} replaces $\ell_2$
	with a learned temporal distance $d_\psi$: a quasimetric head predicts how
	many steps separate two latents, and the planner ranks a candidate by the
	predicted temporal distance from its terminal latent to the goal. TD-JEPA
	blends that distance with $\ell_2$ as
	$(1-\alpha)\,d_\psi + \alpha\,\ell_2$ and compares
	$\alpha \in \{0, 0.1, 1\}$, where $\alpha = 1$ is the plain $\ell_2$
	cost. As published
	the head trains jointly with the encoder and predictor, shaping the
	representation itself. Here, to make the baseline comparable, we fit only
	the head, leaving LeWM's encoder and predictor frozen.
	\item \textbf{PCA} is a standard linear dimensionality reduction
	\citep{jolliffe2002principal}. We take the covariance of the latents,
	project onto its top $k$ eigenvectors, and measure the $\ell_2$ distance between
	the predicted terminal latent and the goal latent. The rank is swept over the
	same grid as our cost, $k \in \{2, 4, 8, \ldots, 128, 192\}$, and selected
	the same way.
	Note that SIGReg
	drives the marginal toward $\mathcal{N}(0, I)$, whose covariance is the
	identity; at the optimum PCA reduces to projecting onto a random
	orthonormal basis.
\end{itemize}

Both run under the planning setup of \Cref{subsec:setup}, on all six
tasks and in both the default and harder settings.

\begin{figure}[t]
	\centering
	\includegraphics[width=\textwidth]{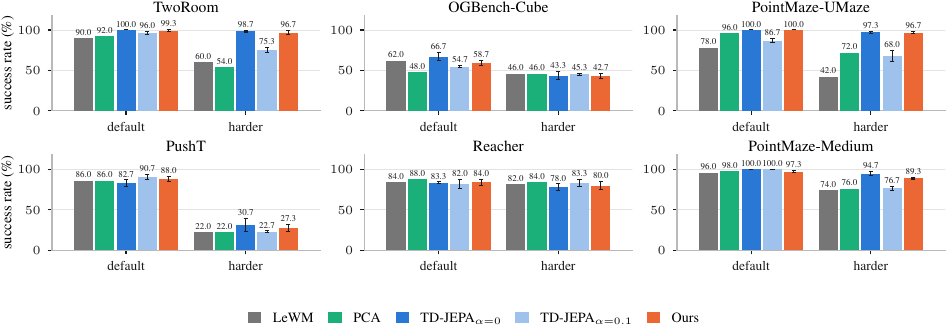}
	\caption{Planning success rate (\%) under five planning costs, evaluated on
	$50$ episodes. TD-JEPA blends its distance with the plain $\ell_2$ cost as
	$(1-\alpha)\,d_\psi + \alpha\,\ell_2$, and both $\alpha = 0$ and
	$\alpha = 0.1$ are reported.
	PCA takes the covariance of the latents and keeps its top
	eigenvectors. PCA and our cost are reported at the retained rank $k^\star$
	selected on the validation seed, swept over
	$k \in \{2, 4, 8, \ldots, 128, 192\}$. Both the state dynamics $A$ and $d_\psi$
	are fit with $3$ seeds; bars are the mean over those seeds, and error bars
	one standard deviation.}
	\label{fig:costbaselines}
\end{figure}

\Cref{fig:costbaselines} reports the comparison on the test seed.
The observation from the validation seed (\Cref{subsec:mainexp}) holds,
with large gains under the harder setting: $+36.7$ on TwoRoom, $+54.7$ on
PointMaze-UMaze and $+15.3$ on PointMaze-Medium over LeWM.
Among the baselines, TD-JEPA trains its
cost with additional supervision, the number of steps separating two latents,
while our cost comes purely from the frozen representation. Despite that, the
two are comparable: averaged over tasks we are $-0.9$ default, $-1.7$ harder,
both against $\alpha = 0$. PCA also measures the cost on a subspace, but chooses it
by variance in the embedding rather than variance in the dynamics, and we
outperform it by a greater margin, $+3.2$ default and $+13.1$ harder.

\subsection{Varying the Latent Regularizer}
\label{subsec:regularizer}

The derivation in \Cref{sec:transition} starts from the latent marginals
being standard Gaussian, which SIGReg supplies in training. To test whether the
latent projection still works when that structure is weakened or removed, we
train
three further world models on TwoRoom with the LeWM recipe, replacing only
the SIGReg term in \eqref{eq:lewmloss}:
\begin{itemize}[leftmargin=*, topsep=2pt, itemsep=2pt]
	\item \textbf{VISReg} \citep{wu2026visreg} targets the same isotropic
	Gaussian, enforced with a sliced-Wasserstein term rather than the
	Epps--Pulley statistic.
	\item \textbf{RDMReg} \citep{kuang2026rectified, kuang2026lpwm} matches
	the latent to a rectified Laplace --- the coordinate-wise
	$\mathrm{ReLU}$ of a Laplace --- yielding a sparse, non-negative code.
	\item \textbf{Stop-grad} \citep{chen2021exploring}
	uses no explicit regularizer at all: collapse is prevented by detaching the
	prediction target, and the marginals are unconstrained. With no
	regularizer weight, we tune the encoder learning rate instead.
\end{itemize}

To compare the gain across variants, we bring planning success under the
plain $\ell_2$ cost close together by modulating the
regularization strength. Each variant then reruns the sweep of
\Cref{subsec:mainexp} under the harder TwoRoom setting and is reported at the
test seed, everything else held fixed. To measure how well the Gaussian assumption actually holds, we
evaluate the Epps--Pulley statistic \citep{epps1983test} of SIGReg
\citep{balestriero2025lejepa} on held-out latents.

\begin{table}[t]
	\caption{Comparison across anti-collapse methods on the harder TwoRoom
		setting. Each variant is a retrained LeWM recipe with only the
		regularizer swapped, at a strength chosen to bring their success
		rate under the plain $\ell_2$ cost close together.
		\emph{Gaussian} marks whether the method targets a Gaussian
		marginal. \emph{Epps--Pulley} measures the Gaussianity of the
		latent. \emph{Gap closed} is the gap that our cost closes,
		$(\mathrm{SR}_{\mathrm{ours}} - \mathrm{SR}_{\mathrm{baseline}})
		/(100 - \mathrm{SR}_{\mathrm{baseline}})$, where $\mathrm{SR}$ is the
		planning success rate, measured under the plain $\ell_2$ cost
		($\mathrm{SR}_{\mathrm{baseline}}$) and under our cost at the best
		retained rank $k^\star$ ($\mathrm{SR}_{\mathrm{ours}}$); both
		appear in parentheses. Latent regularizers that encourage Gaussian
		structure close the largest share of the gap.}
	\centering
	\small
	\setlength{\tabcolsep}{6pt}
	\begin{tabular}{llcc!{\color{black}\vrule}>{\hspace{1em}}c<{\hspace{1em}}}
		\toprule
		Regularizer & Strength & Gaussian & Epps--Pulley $\downarrow$ & Gap closed $\uparrow$ \\
		\midrule
		\rowcolor{mainrow}
		SIGReg (LeWM) & $w = 0.09$ & \checkmark & $2.83$ & $92\%$ {\scriptsize $(60.0 \rightarrow 96.7)$} \\
		VISReg & $w = 20$ & \checkmark & $1.52$ & $98\%$ {\scriptsize $(40.0 \rightarrow 98.7)$} \\
		RDMReg & $w = 20$ & $\times$ & $23.80$ & $87\%$ {\scriptsize $(58.0 \rightarrow 94.7)$} \\
		Stop-grad & $\mathrm{lr} = 3\mathrm{e}{-6}$ & $\times$ & $42.63$ & $0\%$ {\scriptsize $(52.0 \rightarrow 52.0)$} \\
		\bottomrule
	\end{tabular}
	\label{tab:regularizer}
\end{table}

\Cref{tab:regularizer} reports the outcome. The plain-$\ell_2$ scores land
close but not equal, so we report the gain as a fraction of the headroom each variant has
left (\emph{gap closed}). Every
variant with an explicit regularizer term benefits from the projection.
The recovery weakens as the latent moves away from the Gaussian: the
non-Gaussian target of RDMReg gives up part of the gain ($87\%$), and
stop-grad, with no regularizer at all, leaves the latent with no
structure for the projection to exploit ($0\%$). The gain reflects how well the
embedding carries the Gaussian structure assumed in \Cref{sec:transition}.

\subsection{Ablating the Structural Constraints}
\label{subsec:ablation}

The parameterization imposes structural constraints on
$A$, and we ablate each component. (i) \emph{Positive semidefinite}
(PSD): the spectrum is kept below $1$ in magnitude
so that $\Sigma = I - AA^\top$ remains a valid covariance.
(ii) \emph{Normal}: $AA^\top = A^\top A$, so one projection basis serves
every horizon (\Cref{subsec:normality}). (iii) \emph{Motion}: the blocks
are $2 \times 2$ rotation--scaling planes (\Cref{subsec:normality});
with $1 \times 1$ blocks only, $A$ reduces to a symmetric matrix, in which
case $L = n$. (iv) \emph{Multi-step}: $A$ fits $H = 4$ steps
jointly \eqref{eq:multistepfit} rather than the $1$-step transition
alone.

Each variant reruns the sweep of \Cref{subsec:mainexp} on the harder
setting of TwoRoom and PointMaze-UMaze, the two tasks we ablate on, and is
reported at the test seed, everything else held fixed;
\Cref{tab:ablation} reports the result.
The full parameterization attains the best success rate,
and dropping the normality
constraint costs the most on both tasks. The structure drives the gain: an
unstructured linear fit improves on LeWM by only $+22.0$ on
TwoRoom and not at all on PointMaze-UMaze, while the full
parameterization of \Cref{sec:transition} gains $+36.7$ and $+54.7$.

\begin{table}[t]
	\caption{Ablating the structure of $A$ on the harder setting of TwoRoom
	and PointMaze-UMaze. Each row keeps (\checkmark) or removes ($\times$)
	components of the full parameterization (shaded). Success rate is
	measured at the best retained rank $k^\star$, mean $\pm$ std over $3$
	seeds. The last row is LeWM, which plans with the plain $\ell_2$ cost.
	\emph{PSD}: $\Sigma$ remains a valid covariance. \emph{Normal}:
	$AA^\top = A^\top A$. \emph{Motion}: the $2 \times 2$ rotating planes;
	without them $A$ is a symmetric matrix. \emph{Multi-step}: fitting multi-step transitions instead of the
	single step. Every component contributes.}
	\centering
	\small
	\setlength{\tabcolsep}{6pt}
	\begin{tabular}{lcccc!{\color{black}\vrule}cc}
		\toprule
		& PSD & Normal & Motion & Multi-step & TwoRoom & PointMaze-UMaze \\
		\midrule
		\rowcolor{mainrow}
		Ours & \checkmark & \checkmark & \checkmark & \checkmark & \textbf{96.7}\pmsd{2.3} & \textbf{96.7}\pmsd{1.2} \\
		& $\times$ & \checkmark & \checkmark & \checkmark & 96.0\pmsd{2.0} & 94.7\pmsd{1.2} \\
		& \checkmark & $\times$ & --- & \checkmark & 88.0\pmsd{2.0} & 90.7\pmsd{3.1} \\
		& \checkmark & \checkmark & $\times$ & \checkmark & 95.3\pmsd{1.2} & 96.0\pmsd{2.0} \\
		& \checkmark & \checkmark & \checkmark & $\times$ & 94.7\pmsd{5.0} & 92.0\pmsd{5.3} \\
		& $\times$ & $\times$ & --- & \checkmark & 82.0\pmsd{2.0} & 42.0\pmsd{0.0} \\
		\midrule
		LeWM & --- & --- & --- & --- & 60.0 & 42.0 \\
		\bottomrule
	\end{tabular}
	\label{tab:ablation}
\end{table}

\section{Related Work}
\label{sec:related}

\paragraph{Planning in latent space.}
Model-based planning on a learned latent has a long lineage
\citep{ha2018recurrent, schrittwieser2020mastering, hansen2022temporal}. We study model
predictive control (MPC) \citep{richalet1978model, richards2005robust} in
latent space: roll the model forward over candidate action sequences, rank them,
execute, replan. PlaNet \citep{hafner2019learning} does this on a recurrent state-space model,
ranking rollouts by a learned reward. Nearer to us are models that define the
planning cost in representation space: DINO-WM
\citep{zhou2024dinowm} fits a transition on frozen DINO features; V-JEPA~2-AC
\citep{assran2025v} post-trains an action-conditioned predictor on a
frozen video encoder; LeWM \citep{maes2026leworldmodel} trains one end to end
from pixels. They rank candidate sequences by the distance between
the predicted terminal latent and the goal latent, $\ell_2$ in LeWM
and DINO-WM and $\ell_1$ in V-JEPA~2-AC, so every latent direction counts the same. In this work, we show
that the latent holds content irrelevant to planning, that there exists a
clean subspace that carries what matters, and that the distance restricted
to that subspace plans better.

\paragraph{Beyond the plain distance.}
A recent line of work on latent MPC shares one premise: the latent
distance alone is a poor proxy for the reachability the planner needs.
The approaches split in two. One line trains the
representation so that the plain distance is aligned with what planning
needs \citep{zhang2020learning, park2024foundation, wang2026temporal,
thil2026subspace, li2026predictive}. The other redefines the cost, commonly
regressing a distance on trajectory step gaps as a surrogate for
reachability \citep{li2026beyond, bai2026temporal}. We keep the latent
frozen and train no distance function; instead we fit the state dynamics, a
linear function under normality constraints, and compute the cost on the
subspace it determines.

\paragraph{Structure in the latent.}
Self-supervised learning (SSL) yields representations with emergent
properties \citep{chen2020simple, caron2021emerging, radford2021learning,
zhai2023sigmoid, he2022masked, oquab2023dinov2, ranasinghe2022self}. Among these, a JEPA learns by
latent prediction \citep{lecun2022path, assran2023self, bardes2024revisiting,
maes2026leworldmodel}. Without an anti-collapse term the representation collapses
\citep{tian2021understanding, jing2021understanding}. To remedy this, a
stop-gradient with an EMA target was used \citep{grill2020bootstrap, chen2021exploring}. Other methods enforce structure on the embedding itself
through variance and covariance constraints
\citep{bardes2021vicreg, zbontar2021barlow, ermolov2021whitening} and, more
recently, through an isotropic Gaussian target: SIGReg
\citep{balestriero2025lejepa} and subsequent work
\citep{zhao2026sub, wu2026visreg, kuhn2026levjepa,
zimmermann2026kerjepa}. Under this target, \citet{klindt2026does} prove the
learned representation recovers the world's latents up to rotation, a
guarantee unique to the Gaussian. LeWM adopts SIGReg, making its
latent stationary under the transition \citep{maes2026leworldmodel}. But the
regularizer can harm the representation: the latent is intrinsically low-dimensional, and an
isotropic prior on the latent space is an overly strong bias
\citep{zhao2026sub}. We instead use the very structure the regularizer
imposes to identify the subspace with clean state dynamics.

\section{Conclusion}

In this work we show that the state dynamics of the latent select the
subspace a planning cost should measure on. Under the
standard Gaussian marginal SIGReg supplies and a Gauss--Markov model, they are linear and their
covariance is pinned by the map itself,
$z' \mid z \sim \mathcal{N}(Az,\, I - AA^\top)$. Constraining $A$ to be a normal
matrix fixes one eigenbasis across every horizon, and the eigenvalues of
$I - AA^\top$ order that basis by the uncertainty the transition leaves along
each direction. Measuring the planning cost on the subspace spanned by the $k$ most
certain directions delivers large gains on the 2D navigation tasks. The broader point is that the regularizer preventing
collapse both injects the noise and supplies the structure to remove it, and
that exploiting that structure yields a clean planning signal without
supervision.

\paragraph{Limitations and future work.}
We address the cost, but planning can fail in the representation and in the
rollout as well, and restricting the cost to a subspace repairs neither. The derivation also rests
on two modeling choices: a Gauss--Markov latent sequence, and
a normal matrix $A$. Future work will explore training a world model that satisfies
these constraints. Reachability is also a quasimetric, an asymmetric distance
function, while our cost is an $\ell_2$ distance on a subspace and so
symmetric. Our method does fit the state
dynamics, however, and they can be used to compute the state
occupancy measure \citep{sutton1998reinforcement}, a surrogate for
reachability. Learning the state
dynamics can therefore be a building block for approximating reachability,
which we see as an interesting research direction.
\label{sec:conclusion}

\acknowledgments{We thank Varun Belagali, Kanchana Ranasinghe, and Cristina
Mata for helpful discussions and feedback. This research was financially
supported by the Ministry of Trade, Industry, and Energy (MOTIE), Korea,
under the ``Global Industrial Technology Cooperation Center (GITCC)
program'' supervised by the Korea Institute for Advancement of Technology
(KIAT) (Task No.\ P0028420). This research was supported by the National
Research Council of Science \& Technology (NST) grant by the Korea
government (MSIT) (No.\ GTL25041-000). This work was also supported by
the Institute of Information \& Communications Technology Planning \&
Evaluation (IITP) grant funded by the Korea government (MSIT)
(No.\ RS-2024-00336738, Development of Complex Task Planning Technologies
for Autonomous Agents).}

\bibliography{references}

@article{klindt2026does,
  title={When Does LeJEPA Learn a World Model?},
  author={Klindt, David and LeCun, Yann and Balestriero, Randall},
  journal={arXiv preprint arXiv:2605.26379},
  year={2026}
}

@article{wu2026visreg,
  title={VISReg: Variance-Invariance-Sketching Regularization for JEPA training},
  author={Wu, Haiyu and Balestriero, Randall and Levine, Morgan},
  journal={arXiv preprint arXiv:2606.02572},
  year={2026}
}

@article{zhou2024dinowm,
  title={Dino-wm: World models on pre-trained visual features enable zero-shot planning},
  author={Zhou, Gaoyue and Pan, Hengkai and LeCun, Yann and Pinto, Lerrel},
  journal={arXiv preprint arXiv:2411.04983},
  year={2024}
}

@article{balestriero2025lejepa,
  title={{LeJEPA}: Provable and Scalable Self-Supervised Learning Without the Heuristics},
  author={Balestriero, Randall and LeCun, Yann},
  journal={arXiv preprint arXiv:2511.08544},
  year={2025}
}

@article{epps1983test,
  title={A test for normality based on the empirical characteristic function},
  author={Epps, Thomas W and Pulley, Lawrence B},
  journal={Biometrika},
  volume={70},
  number={3},
  pages={723--726},
  year={1983},
  publisher={Oxford University Press}
}

@article{zhang2026identifiability,
  title={On the Identifiability of Controlled World Models},
  author={Zhang, Xiangteng and Guan, Yang and Zhang, Bo and Zhang, Ya-Qin and Li, Shengbo Eben},
  journal={arXiv preprint arXiv:2607.22430},
  year={2026}
}

@article{li2026predictive,
  title={Predictive but Not Plannable: RC-aux for Latent World Models},
  author={Li, Wenyuan and Li, Guang and Maeda, Keisuke and Ogawa, Takahiro and Haseyama, Miki},
  journal={arXiv preprint arXiv:2605.07278},
  year={2026}
}

@article{bai2026temporal,
  title={Temporal-Distance JEPA: Plan-Aware Representation Learning for Latent World Model Predictive Control},
  author={Bai, Jiaxin and Xiong, Jiaxuan},
  journal={arXiv preprint arXiv:2607.25337},
  year={2026}
}

@article{thil2026subspace,
  title={Subspace-Decomposed JEPAs: Disentangling Progression and Content in Latent World Models},
  author={Thil, Lucas and Read, Jesse and Kaddah, Rim and Doquet, Guillaume},
  journal={arXiv preprint arXiv:2605.31111},
  year={2026}
}

@article{maes2026leworldmodel,
  title={Leworldmodel: Stable end-to-end joint-embedding predictive architecture from pixels},
  author={Maes, Lucas and Lidec, Quentin Le and Scieur, Damien and LeCun, Yann and Balestriero, Randall},
  journal={arXiv preprint arXiv:2603.19312},
  year={2026}
}

@article{wang2026temporal,
  title={Temporal straightening for latent planning},
  author={Wang, Ying and Bounou, Oumayma and Zhou, Gaoyue and Balestriero, Randall and Rudner, Tim GJ and LeCun, Yann and Ren, Mengye},
  journal={arXiv preprint arXiv:2603.12231},
  year={2026}
}

@inproceedings{lezcano2019cheap,
  title={Cheap orthogonal constraints in neural networks: A simple parametrization of the orthogonal and unitary group},
  author={Lezcano-Casado, Mario and Mart{\i}nez-Rubio, David},
  booktitle={International conference on machine learning},
  pages={3794--3803},
  year={2019},
  organization={PMLR}
}

@article{zhao2026sub,
  title={Sub-JEPA: Subspace Gaussian Regularization for Stable End-to-End World Models},
  author={Zhao, Kai and Nie, Dongliang and Lin, Yuchen and Luo, Zhehan and Gu, Yixiao and Fan, Deng-Ping and Zeng, Dan},
  journal={arXiv preprint arXiv:2605.09241},
  year={2026}
}

@article{kuhn2026levjepa,
  title={{LeVJEPA}: Efficient \& Scalable Video Pretraining without the Heuristics},
  author={Kuhn, Lukas and Maes, Lucas and Serra, Giuseppe and Le Lidec, Quentin and LeCun, Yann and Balestriero, Randall and Buettner, Florian},
  journal={arXiv preprint arXiv:2608.27395},
  year={2026}
}

@article{li2026beyond,
  title={World Model Control by Trajectory Reachability Metrics},
  author={Li, Liangyu and Wang, Shengzhi and Qiu, Libin and Xiong, Mingliang
          and Liu, Qingwen},
  journal={arXiv preprint arXiv:2605.22164},
  year={2026}
}

@article{lecun2022path,
  title={A path towards autonomous machine intelligence version 0.9. 2, 2022-06-27},
  author={LeCun, Yann and others},
  journal={Open Review},
  volume={62},
  number={1},
  pages={1--62},
  year={2022}
}

@article{kuang2026rectified,
  title={Rectified {LpJEPA}: Joint-Embedding Predictive Architectures with Sparse and Maximum-Entropy Representations},
  author={Kuang, Yilun and Dagade, Yash and Rudner, Tim G. J. and Balestriero, Randall and LeCun, Yann},
  journal={arXiv preprint arXiv:2602.01456},
  year={2026}
}

@book{rubinstein2004cross,
  title={The cross-entropy method: a unified approach to combinatorial optimization, Monte-Carlo simulation, and machine learning},
  author={Rubinstein, Reuven Y and Kroese, Dirk P},
  volume={133},
  year={2004},
  publisher={Springer}
}

@inproceedings{chen2021exploring,
  title={Exploring simple siamese representation learning},
  author={Chen, Xinlei and He, Kaiming},
  booktitle={2021 IEEE/CVF conference on computer vision and pattern recognition (CVPR)},
  pages={15745--15753},
  year={2021},
  organization={IEEE}
}

@article{grill2020bootstrap,
  title={Bootstrap your own latent-a new approach to self-supervised learning},
  author={Grill, Jean-Bastien and Strub, Florian and Altch{\'e}, Florent and Tallec, Corentin and Richemond, Pierre and Buchatskaya, Elena and Doersch, Carl and Avila Pires, Bernardo and Guo, Zhaohan and Gheshlaghi Azar, Mohammad and others},
  journal={Advances in neural information processing systems},
  volume={33},
  pages={21271--21284},
  year={2020}
}

@article{bardes2021vicreg,
  title={Vicreg: Variance-invariance-covariance regularization for self-supervised learning},
  author={Bardes, Adrien and Ponce, Jean and LeCun, Yann},
  journal={arXiv preprint arXiv:2105.04906},
  year={2021}
}

@inproceedings{park2025ogbench,
  title={{OGBench}: Benchmarking offline goal-conditioned {RL}},
  author={Park, Seohong and Frans, Kevin and Eysenbach, Benjamin and Levine, Sergey},
  booktitle={International Conference on Learning Representations},
  year={2025}
}

@article{fu2020d4rl,
  title={{D4RL}: Datasets for deep data-driven reinforcement learning},
  author={Fu, Justin and Kumar, Aviral and Nachum, Ofir and Tucker, George and Levine, Sergey},
  journal={arXiv preprint arXiv:2004.07219},
  year={2020}
}

@software{gymnasium_robotics2023github,
  author = {Rodrigo de Lazcano and Kallinteris Andreas and Jun Jet Tai and Seungjae Ryan Lee and Jordan Terry},
  title = {Gymnasium Robotics},
  url = {http://github.com/Farama-Foundation/Gymnasium-Robotics},
  version = {1.4.0},
  year = {2024},
}

@inproceedings{sobal2025stress,
  title={Stress-testing offline reward-free reinforcement learning: A case for planning with latent dynamics models},
  author={Sobal, Vlad and Zhang, Wancong and Cho, Kyunghyun and Balestriero, Randall and Rudner, Tim G. J. and LeCun, Yann},
  booktitle={7th Robot Learning Workshop: Towards Robots with Human-Level Abilities},
  year={2025}
}

@article{tassa2018deepmind,
  title={{DeepMind} control suite},
  author={Tassa, Yuval and Doron, Yotam and Muldal, Alistair and Erez, Tom and Li, Yazhe and Casas, Diego de Las and Budden, David and Abdolmaleki, Abbas and Merel, Josh and Lefrancq, Andrew and others},
  journal={arXiv preprint arXiv:1801.00690},
  year={2018}
}

@article{kuang2026lpwm,
  title={{LpWM}: A Case for Sparse Representations in World Models},
  author={Kuang, Yilun and Dagade, Yash and Lidec, Quentin Le and Maes, Lucas and Balestriero, Randall and LeCun, Yann},
  journal={arXiv preprint arXiv:2608.22764},
  year={2026}
}

@article{hafner2025mastering,
  title={Mastering diverse control tasks through world models},
  author={Hafner, Danijar and Pasukonis, Jurgis and Ba, Jimmy and Lillicrap, Timothy},
  journal={Nature},
  volume={640},
  number={8059},
  pages={647--653},
  year={2025},
  publisher={Nature Publishing Group UK London}
}

@inproceedings{hansen2024td,
  title={Td-mpc2: Scalable, robust world models for continuous control},
  author={Hansen, Nick and Su, Hao and Wang, Xiaolong},
  booktitle={International Conference on Learning Representations},
  volume={2024},
  pages={47376--47405},
  year={2024}
}

@inproceedings{hafner2019learning,
  title={Learning latent dynamics for planning from pixels},
  author={Hafner, Danijar and Lillicrap, Timothy and Fischer, Ian and Villegas, Ruben and Ha, David and Lee, Honglak and Davidson, James},
  booktitle={International conference on machine learning},
  pages={2555--2565},
  year={2019},
  organization={PMLR}
}

@book{horn2012matrix,
  title={Matrix analysis},
  author={Horn, Roger A and Johnson, Charles R},
  year={2012},
  publisher={Cambridge university press}
}

@software{minari,
	author = {Younis, Omar G. and Perez-Vicente, Rodrigo and Balis, John U. and Dudley, Will and Davey, Alex and Terry, Jordan K},
	doi = {10.5281/zenodo.13767625},
	month = sep,
	publisher = {Zenodo},
	title = {Minari},
	url = {https://doi.org/10.5281/zenodo.13767625},
	version = {0.5.0},
	year = 2024
}

@article{chi2025diffusion,
	author = {Cheng Chi and Zhenjia Xu and Siyuan Feng and Eric Cousineau and Yilun Du and Benjamin Burchfiel and Russ Tedrake and Shuran Song},
	title = {Diffusion Policy: Visuomotor Policy Learning via Action Diffusion},
	journal = {The International Journal of Robotics Research},
	year = {2025}
}

@book{jolliffe2002principal,
  title={Principal component analysis},
  author={Jolliffe, Ian T},
  year={2002},
  publisher={Springer}
}

@article{ha2018recurrent,
  title={Recurrent world models facilitate policy evolution},
  author={Ha, David and Schmidhuber, J{\"u}rgen},
  journal={Advances in neural information processing systems},
  volume={31},
  year={2018}
}

@article{schrittwieser2020mastering,
  title={Mastering atari, go, chess and shogi by planning with a learned model},
  author={Schrittwieser, Julian and Antonoglou, Ioannis and Hubert, Thomas and Simonyan, Karen and Sifre, Laurent and Schmitt, Simon and Guez, Arthur and Lockhart, Edward and Hassabis, Demis and Graepel, Thore and others},
  journal={Nature},
  volume={588},
  number={7839},
  pages={604--609},
  year={2020},
  publisher={Nature Publishing Group UK London}
}

@article{hansen2022temporal,
  title={Temporal difference learning for model predictive control},
  author={Hansen, Nicklas and Wang, Xiaolong and Su, Hao},
  journal={arXiv preprint arXiv:2203.04955},
  year={2022}
}

@article{richalet1978model,
  title={Model predictive heuristic control},
  author={Richalet, Jacques and Rault, Andr{\'e} and Testud, Jean-Louis and Papon, Jean},
  journal={Automatica (journal of IFAC)},
  volume={14},
  number={5},
  pages={413--428},
  year={1978},
  publisher={Pergamon Press, Inc. Elmsford, NY, USA}
}

@inproceedings{richards2005robust,
  title={Robust model predictive control with imperfect information},
  author={Richards, Arthur and How, Jonathan},
  booktitle={Proceedings of the 2005, American Control Conference, 2005.},
  pages={268--273},
  year={2005},
  organization={IEEE}
}

@article{assran2025v,
  title={V-jepa 2: Self-supervised video models enable understanding, prediction and planning},
  author={Assran, Mido and Bardes, Adrien and Fan, David and Garrido, Quentin and Howes, Russell and Muckley, Matthew and Rizvi, Ammar and Roberts, Claire and Sinha, Koustuv and Zholus, Artem and others},
  journal={arXiv preprint arXiv:2506.09985},
  year={2025}
}

@article{zhang2020learning,
  title={Learning invariant representations for reinforcement learning without reconstruction},
  author={Zhang, Amy and McAllister, Rowan and Calandra, Roberto and Gal, Yarin and Levine, Sergey},
  journal={arXiv preprint arXiv:2006.10742},
  year={2020}
}

@article{park2024foundation,
  title={Foundation policies with hilbert representations},
  author={Park, Seohong and Kreiman, Tobias and Levine, Sergey},
  journal={arXiv preprint arXiv:2402.15567},
  year={2024}
}

@inproceedings{chen2020simple,
  title={A simple framework for contrastive learning of visual representations},
  author={Chen, Ting and Kornblith, Simon and Norouzi, Mohammad and Hinton, Geoffrey},
  booktitle={International conference on machine learning},
  pages={1597--1607},
  year={2020},
  organization={PmLR}
}

@inproceedings{caron2021emerging,
  title={Emerging properties in self-supervised vision transformers},
  author={Caron, Mathilde and Touvron, Hugo and Misra, Ishan and J{\'e}gou, Herv{\'e} and Mairal, Julien and Bojanowski, Piotr and Joulin, Armand},
  booktitle={2021 IEEE/CVF international conference on computer vision (ICCV)},
  pages={9630--9640},
  year={2021},
  organization={IEEE}
}

@inproceedings{radford2021learning,
  title={Learning transferable visual models from natural language supervision},
  author={Radford, Alec and Kim, Jong Wook and Hallacy, Chris and Ramesh, Aditya and Goh, Gabriel and Agarwal, Sandhini and Sastry, Girish and Askell, Amanda and Mishkin, Pamela and Clark, Jack and others},
  booktitle={International conference on machine learning},
  pages={8748--8763},
  year={2021},
  organization={PmLR}
}

@inproceedings{ranasinghe2022self,
  title={Self-supervised video transformer},
  author={Ranasinghe, Kanchana and Naseer, Muzammal and Khan, Salman and Khan, Fahad Shahbaz and Ryoo, Michael S},
  booktitle={2022 IEEE/CVF Conference on Computer Vision and Pattern Recognition (CVPR)},
  pages={2864--2874},
  year={2022},
  organization={IEEE}
}

@inproceedings{zhai2023sigmoid,
  title={Sigmoid loss for language image pre-training},
  author={Zhai, Xiaohua and Mustafa, Basil and Kolesnikov, Alexander and Beyer, Lucas},
  booktitle={2023 IEEE/CVF International Conference on Computer Vision (ICCV)},
  pages={11941--11952},
  year={2023},
  organization={IEEE}
}

@inproceedings{he2022masked,
  title={Masked autoencoders are scalable vision learners},
  author={He, Kaiming and Chen, Xinlei and Xie, Saining and Li, Yanghao and Doll{\'a}r, Piotr and Girshick, Ross},
  booktitle={2022 IEEE/CVF conference on computer vision and pattern recognition (CVPR)},
  pages={15979--15988},
  year={2022},
  organization={IEEE}
}

@article{oquab2023dinov2,
  title={Dinov2: Learning robust visual features without supervision},
  author={Oquab, Maxime and Darcet, Timoth{\'e}e and Moutakanni, Th{\'e}o and Vo, Huy and Szafraniec, Marc and Khalidov, Vasil and Fernandez, Pierre and Haziza, Daniel and Massa, Francisco and El-Nouby, Alaaeldin and others},
  journal={arXiv preprint arXiv:2304.07193},
  year={2023}
}

@inproceedings{assran2023self,
  title={Self-supervised learning from images with a joint-embedding predictive architecture},
  author={Assran, Mahmoud and Duval, Quentin and Misra, Ishan and Bojanowski, Piotr and Vincent, Pascal and Rabbat, Michael and LeCun, Yann and Ballas, Nicolas},
  booktitle={2023 IEEE/CVF Conference on Computer Vision and Pattern Recognition (CVPR)},
  pages={15619--15629},
  year={2023},
  organization={IEEE}
}

@article{bardes2024revisiting,
  title={Revisiting feature prediction for learning visual representations from video},
  author={Bardes, Adrien and Garrido, Quentin and Ponce, Jean and Chen, Xinlei and Rabbat, Michael and LeCun, Yann and Assran, Mahmoud and Ballas, Nicolas},
  journal={arXiv preprint arXiv:2404.08471},
  year={2024}
}

@inproceedings{tian2021understanding,
  title={Understanding self-supervised learning dynamics without contrastive pairs},
  author={Tian, Yuandong and Chen, Xinlei and Ganguli, Surya},
  booktitle={International Conference on Machine Learning},
  pages={10268--10278},
  year={2021},
  organization={PMLR}
}

@article{jing2021understanding,
  title={Understanding dimensional collapse in contrastive self-supervised learning},
  author={Jing, Li and Vincent, Pascal and LeCun, Yann and Tian, Yuandong},
  journal={arXiv preprint arXiv:2110.09348},
  year={2021}
}

@inproceedings{zbontar2021barlow,
  title={Barlow twins: Self-supervised learning via redundancy reduction},
  author={Zbontar, Jure and Jing, Li and Misra, Ishan and LeCun, Yann and Deny, St{\'e}phane},
  booktitle={International conference on machine learning},
  pages={12310--12320},
  year={2021},
  organization={PMLR}
}

@inproceedings{ermolov2021whitening,
  title={Whitening for self-supervised representation learning},
  author={Ermolov, Aleksandr and Siarohin, Aliaksandr and Sangineto, Enver and Sebe, Nicu},
  booktitle={International conference on machine learning},
  pages={3015--3024},
  year={2021},
  organization={PMLR}
}

@article{zimmermann2026kerjepa,
  title={{KerJEPA}: Kernel Discrepancies for {E}uclidean Self-Supervised Learning},
  author={Zimmermann, Eric and Wiltzer, Harley and Szeto, Justin and Alvarez-Melis, David and Mackey, Lester},
  journal={arXiv preprint arXiv:2512.19605},
  year={2026}
}

@inproceedings{haarnoja2018soft,
  title={Soft actor-critic: Off-policy maximum entropy deep reinforcement learning with a stochastic actor},
  author={Haarnoja, Tuomas and Zhou, Aurick and Abbeel, Pieter and Levine, Sergey},
  booktitle={International conference on machine learning},
  pages={1861--1870},
  year={2018},
  organization={Pmlr}
}

@book{sutton1998reinforcement,
  title={Reinforcement learning: An introduction},
  author={Sutton, Richard S and Barto, Andrew G and Barto, Andrew},
  volume={1},
  number={1},
  year={1998},
  publisher={MIT press Cambridge}
}

\clearpage
\appendix

\section{Why Does Normality Matter?}
\label{app:whynormal}
The cost of \Cref{subsec:project} keeps the $k$ most certain directions of
the one-step covariance $\Sigma = I - AA^\top$ and measures the distance
there. The subspace is chosen once, before planning begins, and applied to
every candidate the planner scores. The choice hides a dependence on the
horizon. Over $t$ steps the transition carries noise $I - A^t (A^t)^\top$
\eqref{eq:multistep}, and for a general $A$ the subspace moves with $t$.
No candidate settles the choice of
$t$ --- its remaining gap to the goal is neither given nor fixed. Normality
removes the dependence. Every horizon shares one basis and one ordering
(\Cref{subsec:normality}), so the subspace from the one-step transition
serves them all.

\paragraph{From distance to reachability.} The cost scores the predicted
terminal latent against the goal, as a distance. A reachability cost --- the
state occupancy measure --- accumulates likelihood over horizons instead.
Under normality every horizon is read in one basis, so the sum over $t$
factorizes and amortizes, which could be helpful in approximating
reachability. We leave that to future work.

\vspace{2em}

\section{Parameterization of the State Dynamics}
\label{app:param}

The fit of \Cref{subsec:fit} must respect two constraints: $A$ must be a normal
matrix and $\Sigma = I - A A^\top$ must be a valid covariance,
$\Sigma \succeq 0$ (\Cref{sec:transition}). We parameterize $A$ so that both
hold by construction during the fit.

We take the form from the decomposition $A = Q B Q^\top$ of
\Cref{subsec:normality}. Splitting $Q$ and $B$ between the $L = n - 2P$ lines
and the $P$ planes,
\begin{equation}
	Q = \begin{bmatrix} Q_L & Q_P \end{bmatrix},
	\qquad
	B = \begin{bmatrix} B_L & 0 \\ 0 & B_P \end{bmatrix},
	\label{eq:qbsplit}
\end{equation}
we parameterize each type of block separately,
\begin{align}
	B_L &= \mathrm{diag}(\lambda_\ell)_{\ell=1}^{L},
	\qquad
	\lambda_\ell = \frac{v_\ell}{\sqrt{1 + v_\ell^2}} \in (-1, 1),
	\nonumber \\
	B_P &= \mathrm{blkdiag}\big(r_p R(\theta_p)\big)_{p=1}^{P},
	\qquad
	r_p = \frac{|u_p|}{\sqrt{1 + u_p^2}} \in [0, 1),
	\nonumber \\
	A &= Q_L B_L Q_L^\top + Q_P B_P Q_P^\top ,
	\label{eq:param}
\end{align}
where $Q$ is the matrix exponential of a free skew-symmetric matrix
\citep{lezcano2019cheap}, orthogonal by construction. The free parameters are
that skew-symmetric matrix, the $L$ scalars $v_\ell$, and the $P$ scalars
$u_p$ and angles $\theta_p$.

\paragraph{Normality holds by construction.} Each $2 \times 2$ block is a
scaled rotation, and $R(\theta)$ is orthogonal, so
$(r R)(r R)^\top = (r R)^\top (r R) = r^2 I_2$. Each $1 \times 1$ block is
trivially normal. $B$ is block diagonal with these blocks, so
$B B^\top = B^\top B$, and
\begin{equation}
	A A^\top = Q B B^\top Q^\top = Q B^\top B Q^\top = A^\top A .
\end{equation}

\paragraph{The covariance is positive definite.} The same computation gives
\[
	B B^\top = \mathrm{diag}\big(\lambda_1^2, \ldots, \lambda_L^2,
	r_1^2, r_1^2, \ldots, r_P^2, r_P^2\big),
\]
so
\begin{equation}
	\Sigma = I - A A^\top = Q \big(I - B B^\top\big) Q^\top
	\label{eq:sigmaparam}
\end{equation}
has eigenvalues $1 - \lambda_\ell^2$ and $1 - r_p^2$, the latter appearing twice.
The squashing in \eqref{eq:param} keeps $r_p < 1$ and
$|\lambda_\ell| < 1$, so
every eigenvalue is strictly positive and $\Sigma \succ 0$.

\section{Environments and Datasets}
\label{app:envs}

\begin{figure}[t]
	\centering
	\includegraphics[width=\linewidth]{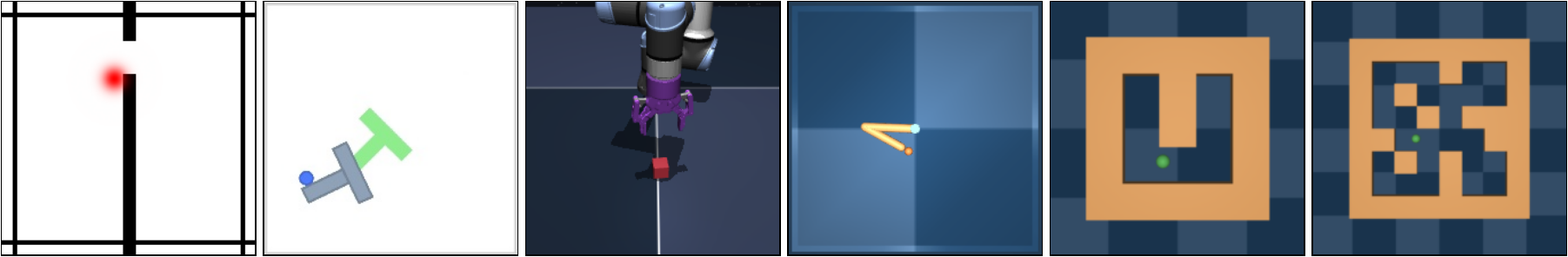}\\[2pt]
	\makebox[\dimexpr\linewidth/6\relax][c]{\small TwoRoom}%
	\makebox[\dimexpr\linewidth/6\relax][c]{\small PushT}%
	\makebox[\dimexpr\linewidth/6\relax][c]{\small OGBench-Cube}%
	\makebox[\dimexpr\linewidth/6\relax][c]{\small Reacher}%
	\makebox[\dimexpr\linewidth/6\relax][c]{\small PointMaze-UMaze}%
	\makebox[\dimexpr\linewidth/6\relax][c]{\small PointMaze-Medium}%
	\caption{The six evaluation tasks.}
	\label{fig:tasks}
\end{figure}

The evaluation covers six tasks. Four of them --- TwoRoom, PushT,
OGBench-Cube and Reacher --- come from LeWM \citep{maes2026leworldmodel}: the
same environments, LeWM's trajectory data, and the released checkpoints, so
the encoder and predictor are frozen throughout and are not retrained here.
For PointMaze-UMaze and PointMaze-Medium, we collect the training data,
following DINO-WM's setting \citep{zhou2024dinowm}, and train the world model
with the LeWM recipe. Every world model is trained for $10$ epochs. Dataset
sizes are given in \Cref{tab:datasets}. Planning succeeds when the state
reached comes within a tolerance of the goal; what that is measured on differs
by task.

\noindent\textbf{TwoRoom.} A simple continuous navigation task in 2D
\citep{sobal2025stress}. The environment has two rooms joined by a single
opening, and wall (black) blocks the agent everywhere else. The agent (red)
navigates to a randomly selected location in the other room, so it
must pass through the opening. The trajectories come from a noisy heuristic
that steers to the opening first and to the target once through it.

\noindent\textbf{PushT.} A contact-rich 2D manipulation task introduced by
\citet{chi2025diffusion}. The agent (blue) pushes a T-shaped block (gray) onto
a T-shaped goal (green). The agent and block positions are scored together,
along with the block's angle up to its rotational symmetry. The data is the
PushT set released with DINO-WM \citep{zhou2024dinowm}, which LeWM adopts.

\noindent\textbf{OGBench-Cube.} A 3D manipulation task from the OGBench suite
\citep{park2025ogbench}. A robot arm moves a cube to the goal position. We use
the single-cube variant. An episode succeeds when the cube ends near the
goal, wherever the arm is. The trajectories come from the scripted policy the benchmark provides
for data collection.

\noindent\textbf{Reacher.} A continuous control task from the DeepMind Control Suite
\citep{tassa2018deepmind}. A two-joint arm reaches for a target. Every joint
angle is scored, so the whole arm has to match and not just the fingertip. The
trajectories come from an expert SAC policy \citep{haarnoja2018soft}.

\noindent\textbf{PointMaze (UMaze and Medium).} A 2D navigation task with
physics \citep{fu2020d4rl}, in two of its maze layouts, run through the
Gymnasium-Robotics implementation \citep{gymnasium_robotics2023github}. A point
mass moves through a walled maze, and the action is a force along $x$ and $y$.
Only the position of the mass is scored. We collect the training data ourselves, rolling out random
action sequences as DINO-WM does \citep{zhou2024dinowm}.

The evaluation start--goal pairs do not come from that random data. On random
rollouts the mass often gets stuck at a corner of the maze, so the two states
of a pair sit close together; a success rate measured that way comes out
inflated. We take the pairs from the D4RL expert
trajectories instead, distributed through Minari \citep{minari}, where the
point mass does cross the maze.
\begin{table}[h]
	\caption{Training data for each world model.
		Lengths marked ``avg.'' are averages over variable-length
		episodes.}
	\centering
	\small
	\begin{tabular}{lrc}
		\toprule
		Task & Episodes & Episode length \\
		\midrule
		TwoRoom            & $10{,}000$ & $92$ (avg.)  \\
		PushT              & $18{,}685$ & $125$ (avg.) \\
		OGBench-Cube       & $10{,}000$ & $200$        \\
		Reacher            & $10{,}000$ & $200$        \\
		PointMaze-UMaze    & $10{,}000$ & $91$ (avg.)  \\
		PointMaze-Medium   & $10{,}000$ & $98$ (avg.)  \\
		\bottomrule
	\end{tabular}
	\label{tab:datasets}
\end{table}

\section{Planning Configuration}
\label{app:planning}

We adopt the planning configuration of LeWM. At each planning step, the CEM
solver draws $N$ candidate action sequences of horizon $T$, scores each by
the planning cost, and retains the best $E$ to update the sampling
distribution; this repeats for $M$ iterations, after which the whole
optimized sequence is executed before replanning. One predictor step covers
$c$ environment timesteps (the frame skip). At evaluation the goal is drawn
$\Delta$ timesteps ahead of the initial state within the same trajectory, and the
agent is allowed a budget of $b$ actions in the environment.
\Cref{tab:planconfig} gives the values for the default and harder settings
of \Cref{subsec:setup}.

\begin{table}[h]
	\caption{Planning configuration for the default and harder settings.}
	\centering
	\begin{tabular}{llcc}
		\toprule
		Symbol & Quantity & Default & Harder \\
		\midrule
		$N$ & candidate sequences per iteration & $300$ & $300$ \\
		$M$ & CEM iterations & $10$ ($30$ on PushT) & $10$ ($30$ on PushT) \\
		$E$ & retained elites & $30$ & $30$ \\
		$T$ & planning horizon (predictor steps) & $5$ & $10$ \\
		$c$ & frame skip (env.\ timesteps per predictor step) & $5$ & $5$ \\
		$\Delta$ & goal offset (timesteps) & $25$ & $50$ \\
		$b$ & evaluation budget (actions) & $50$ & $100$ \\
		\bottomrule
	\end{tabular}
	\label{tab:planconfig}
\end{table}

\section{The Spectrum of the Transition Noise}
\label{app:spectrum}

\begin{figure}[b]
	\centering
	\includegraphics[width=0.8\textwidth]{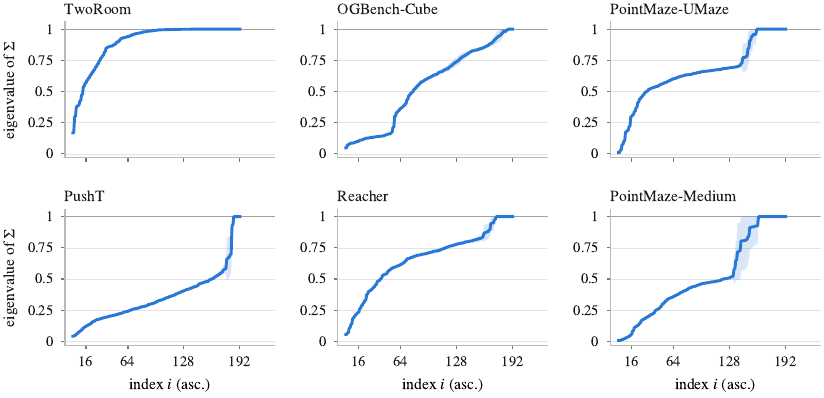}
	\caption{Eigenvalues of $\Sigma = I - AA^\top$ per task, sorted in
	increasing order. Mean $\pm$ std over the $3$ seeds $A$ is fit with.}
	\label{fig:eigval}
\end{figure}

\Cref{fig:eigval} plots the eigenvalues of $\Sigma = I - A A^\top$ for the $A$
fitted to each task, sorted ascending. The eigenvalue
$\sigma_i^2$ is the variance the one-step transition leaves along direction
$u_i$: at $\sigma_i^2 = 0$ the next value is determined, and at
$\sigma_i^2 = 1$ the transition law along $u_i$ is the marginal itself, so the
dynamics carry nothing forward there.

Every spectrum reaches that ceiling. On TwoRoom, PointMaze-UMaze, Reacher and
OGBench-Cube the transition variance is nearer the ceiling than zero over most
of the basis, so the state dynamics carry little forward there.
PushT stands apart: its spectrum stays low across the greater part of the
range.

\section{State Recovery in the Retained Subspace}
\label{app:staterec}

We assume the latent holds the information planning needs. The ground-truth
state of the environment can serve as a proxy for that information, so a
linear probe from the retained directions to the state tests whether they
carry it. We pair cached latents $z \in \mathbb{R}^{192}$ with the
ground-truth state $s$. We keep the $k$ most certain directions $U_k$ of the
fitted eigenbasis --- the same basis the cost projects onto --- and fit
ordinary least squares $\hat{s} = W (U_k^\top z)$ on $10$k frames, then report
held-out $R^2$ on $10$k frames from disjoint episodes. As a control, we run the same
probe on a random orthonormal subspace of rank $k$.

\Cref{tab:staterec} reports the probe. As the rank grows the retained
directions recover the state first, while a random subspace of the same rank
lags well behind. What carries the state is the ordering, then, and not the
number of dimensions. How quickly the state arrives also tracks the rank
planning needs (\Cref{subsec:mainexp}). PushT and OGBench-Cube are the two
tasks slowest to recover, in different ways. PushT climbs gradually, every
coordinate improving with $k$ and none of them arriving early. OGBench-Cube
recovers unevenly instead: the planar coordinates arrive at the very lowest
ranks, while the two heights and the gripper stay absent until well into the
sweep.

\begin{table}[h]
	\centering
	\small
	\setlength{\tabcolsep}{4.2pt}
	\caption{Held-out $R^2$ of a linear probe reading the ground-truth state
	from a $k$-dimensional subspace of the frozen $192$-dimensional latent,
	fit by ordinary least squares on $10$k frames and scored on $10$k
	held-out frames from disjoint episodes. \emph{Ours} keeps the $k$ least-noisy
	eigenvectors of $\Sigma = I - AA^\top$, the directions the cost
	\eqref{eq:projcost} projects onto; \emph{random} is a random orthonormal
	$k$-subspace. Rank
	increases left to right within each group.
	Each \emph{mean} averages the
	rows of its task. Only positions are shown; velocities are omitted.}
	\begin{tabular}{ll cccc cccc c}
		\toprule
		& & \multicolumn{4}{c}{ours} & \multicolumn{4}{c}{random} & \\
		\cmidrule(lr){3-6}\cmidrule(lr){7-10}
		Task & State & $2$ & $8$ & $32$ & $128$ & $2$ & $8$ & $32$ & $128$
		  & $192$ \\
		\midrule
		\multirow{3}{*}{TwoRoom}
		  & $x$ & $0.71$ & $0.94$ & $0.97$ & $0.99$ & $0.03$ & $0.09$ & $0.35$ & $0.99$ & $1.00$ \\
		  & $y$ & $0.23$ & $0.72$ & $0.90$ & $0.98$ & $0.01$ & $0.06$ & $0.29$ & $0.94$ & $0.99$ \\
		\cmidrule(l){2-11}
		  & \emph{mean} & $0.47$ & $0.83$ & $0.94$ & $0.98$ & $0.02$ & $0.07$ & $0.32$ & $0.96$ & $0.99$ \\
		\midrule
		\multirow{7}{*}{PushT}
		  & agent $x$ & $0.02$ & $0.20$ & $0.47$ & $0.93$ & $0.03$ & $0.09$ & $0.32$ & $0.93$ & $0.96$ \\
		  & agent $y$ & $0.03$ & $0.19$ & $0.41$ & $0.92$ & $0.02$ & $0.07$ & $0.33$ & $0.93$ & $0.96$ \\
		  & block $x$ & $0.02$ & $0.36$ & $0.86$ & $0.97$ & $0.03$ & $0.09$ & $0.41$ & $0.97$ & $0.98$ \\
		  & block $y$ & $0.06$ & $0.41$ & $0.71$ & $0.96$ & $0.02$ & $0.07$ & $0.33$ & $0.95$ & $0.97$ \\
		  & $\sin\theta_b$ & $0.01$ & $0.17$ & $0.56$ & $0.84$ & $0.01$ & $0.06$ & $0.30$ & $0.84$ & $0.91$ \\
		  & $\cos\theta_b$ & $0.05$ & $0.23$ & $0.59$ & $0.89$ & $0.01$ & $0.07$ & $0.33$ & $0.88$ & $0.93$ \\
		\cmidrule(l){2-11}
		  & \emph{mean} & $0.03$ & $0.26$ & $0.60$ & $0.92$ & $0.02$ & $0.07$ & $0.34$ & $0.92$ & $0.95$ \\
		\midrule
		\multirow{8}{*}{OGBench-Cube}
		  & block $x$ & $0.92$ & $0.94$ & $0.96$ & $0.99$ & $0.01$ & $0.06$ & $0.31$ & $0.98$ & $0.99$ \\
		  & block $y$ & $0.95$ & $0.96$ & $0.97$ & $0.99$ & $0.01$ & $0.07$ & $0.23$ & $0.98$ & $0.99$ \\
		  & block $z$ & $0.00$ & $0.00$ & $0.00$ & $0.95$ & $0.02$ & $0.06$ & $0.24$ & $0.97$ & $0.99$ \\
		  & effector $x$ & $0.81$ & $0.83$ & $0.84$ & $0.96$ & $0.01$ & $0.06$ & $0.29$ & $0.97$ & $0.99$ \\
		  & effector $y$ & $0.90$ & $0.91$ & $0.92$ & $0.96$ & $0.01$ & $0.07$ & $0.23$ & $0.95$ & $0.97$ \\
		  & effector $z$ & $0.00$ & $0.00$ & $0.00$ & $0.95$ & $0.02$ & $0.06$ & $0.24$ & $0.95$ & $0.98$ \\
		  & gripper & $0.00$ & $0.00$ & $0.00$ & $0.91$ & $0.02$ & $0.08$ & $0.27$ & $0.90$ & $0.92$ \\
		\cmidrule(l){2-11}
		  & \emph{mean} & $0.51$ & $0.52$ & $0.53$ & $0.96$ & $0.01$ & $0.07$ & $0.26$ & $0.96$ & $0.98$ \\
		\midrule
		\multirow{5}{*}{Reacher}
		  & $\sin\theta_0$ & $0.46$ & $0.99$ & $0.99$ & $1.00$ & $0.07$ & $0.10$ & $0.48$ & $1.00$ & $1.00$ \\
		  & $\cos\theta_0$ & $0.67$ & $0.99$ & $0.99$ & $1.00$ & $0.04$ & $0.13$ & $0.48$ & $1.00$ & $1.00$ \\
		  & $\sin\theta_1$ & $0.83$ & $0.95$ & $0.99$ & $1.00$ & $0.03$ & $0.12$ & $0.47$ & $1.00$ & $1.00$ \\
		  & $\cos\theta_1$ & $0.00$ & $0.99$ & $0.99$ & $1.00$ & $0.03$ & $0.11$ & $0.43$ & $1.00$ & $1.00$ \\
		\cmidrule(l){2-11}
		  & \emph{mean} & $0.49$ & $0.98$ & $0.99$ & $1.00$ & $0.04$ & $0.11$ & $0.46$ & $1.00$ & $1.00$ \\
		\midrule
		\multirow{3}{*}{PointMaze-UMaze}
		  & $x$ & $0.96$ & $0.99$ & $1.00$ & $1.00$ & $0.05$ & $0.18$ & $0.69$ & $1.00$ & $1.00$ \\
		  & $y$ & $0.78$ & $0.99$ & $1.00$ & $1.00$ & $0.04$ & $0.22$ & $0.70$ & $1.00$ & $1.00$ \\
		\cmidrule(l){2-11}
		  & \emph{mean} & $0.87$ & $0.99$ & $1.00$ & $1.00$ & $0.04$ & $0.20$ & $0.69$ & $1.00$ & $1.00$ \\
		\midrule
		\multirow{3}{*}{PointMaze-Medium}
		  & $x$ & $0.25$ & $0.91$ & $0.99$ & $1.00$ & $0.03$ & $0.18$ & $0.47$ & $0.99$ & $1.00$ \\
		  & $y$ & $0.31$ & $0.88$ & $0.98$ & $1.00$ & $0.02$ & $0.10$ & $0.49$ & $1.00$ & $1.00$ \\
		\cmidrule(l){2-11}
		  & \emph{mean} & $0.28$ & $0.90$ & $0.98$ & $1.00$ & $0.03$ & $0.14$ & $0.48$ & $0.99$ & $1.00$ \\
		\bottomrule
	\end{tabular}
	\label{tab:staterec}
\end{table}

\end{document}